\documentclass[letterpaper]{article} 
\usepackage{aaai23}  
\usepackage{times}  
\usepackage{helvet}  
\usepackage{courier}  
\usepackage[hyphens]{url}  
\usepackage{graphicx} 
\usepackage{natbib}  
\usepackage{caption} 
\usepackage{algorithm}
\usepackage{algorithmic}
\usepackage{bm}
\usepackage{amsmath}
\usepackage{booktabs}
\usepackage{makecell}
\usepackage{amssymb}
\newcommand{\tabincell}[2]{\begin{tabular}{@{}#1@{}}#2\end{tabular}}

\usepackage{newfloat}
\usepackage{listings}
\DeclareCaptionStyle{ruled}{labelfont=normalfont,labelsep=colon,strut=off} 
\floatstyle{ruled}
\newfloat{listing}{tb}{lst}{}
\floatname{listing}{Listing}
\def \Ours {KVPFormer}

\title{A Question-Answering Approach to Key Value Pair Extraction from Form-like Document Images}
\author{
    Kai Hu\textsuperscript{\rm 1,\rm 3,\thanks{This work was done when Kai Hu and Zhuoyuan Wu were interns in Multi-Modal Interaction Group, Microsoft Research Asia, Beijing, China.},\equalcontrib},
    Zhuoyuan Wu\textsuperscript{\rm 2,\rm 3,\footnotemark[1],\equalcontrib},
    Zhuoyao Zhong\textsuperscript{\rm 3,\thanks{Corresponding author.}},
    Weihong Lin\textsuperscript{\rm 3},
    Lei Sun\textsuperscript{\rm 3},
    Qiang Huo\textsuperscript{\rm 3} \\
}

\affiliations{
    \textsuperscript{\rm 1} University of Science and Technology of China, Hefei, China \\
    \textsuperscript{\rm 2} Peking University Shenzhen Graduate School, Shenzhen, China \\
    \textsuperscript{\rm 3} Microsoft Research Asia, Beijing, China \\
    hk970213@mail.ustc.edu.cn, wuzhuoyuan@pku.edu.cn, zhuoyao.zhong@gmail.com, \\
    lwher1996@gmail.com, kuangtongustc@gmail.com, qianghuo@microsoft.com

}

\usepackage{bibentry}

\begin{document}

\maketitle

\begin{abstract}
In this paper, we present a new question-answering (QA) based key-value pair extraction approach, called KVPFormer, to robustly extracting key-value relationships between entities from form-like document images. Specifically, KVPFormer first identifies key entities from all entities in an image with a Transformer encoder, then takes these key entities as \textbf{questions} and feeds them into a Transformer decoder to predict their corresponding \textbf{answers} (i.e., value entities) in parallel. To achieve higher answer prediction accuracy, we propose a coarse-to-fine answer prediction approach further, which first extracts multiple answer candidates for each identified question in the coarse stage and then selects the most likely one among these candidates in the fine stage. In this way, the learning difficulty of answer prediction can be effectively reduced so that the prediction accuracy can be improved. Moreover, we introduce a spatial compatibility attention bias into the self-attention/cross-attention mechanism for \Ours{} to better model the spatial interactions between entities. With these new techniques, our proposed \Ours{} achieves state-of-the-art results on FUNSD and XFUND datasets, outperforming the previous best-performing method by 7.2\% and 13.2\% in F1 score, respectively.
\end{abstract}

\section{Introduction}

\begin{figure}[!t]
    \centering 
    \includegraphics[width=\linewidth]{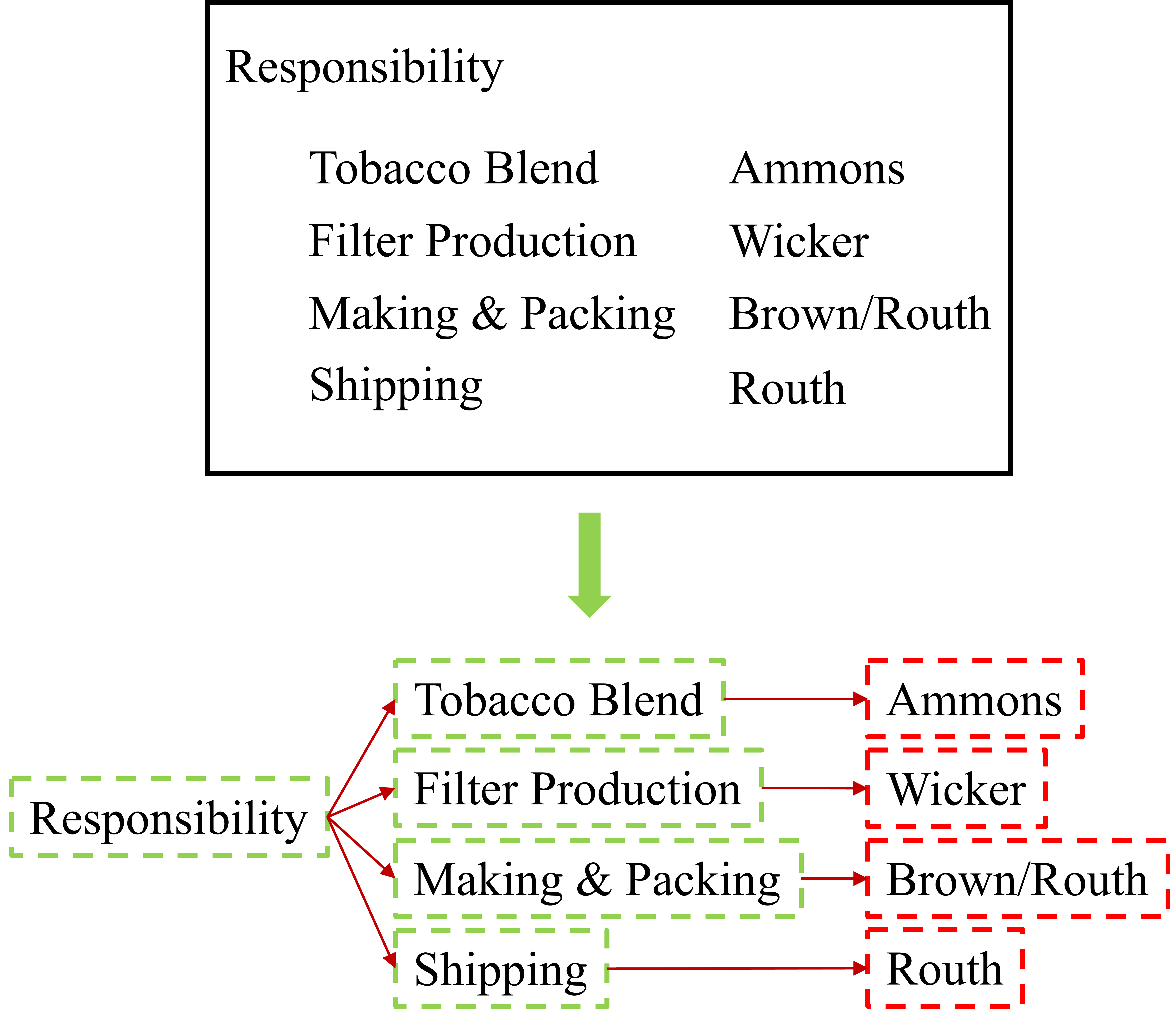}
    \caption{Examples of key-value pairs in the form-like document images. A red arrow stands for a directed key-value relationship pointing from a key entity (green dashed box) to a value entity (red dashed box).}
    \label{FUNSD_showcase}
\end{figure}

Form-like documents like invoices, receipts, purchase orders and tax forms are widely used in day-to-day business workflows. Most of these documents contain critical information represented in the structure of key-value pairs (KVPs). Essentially, a key-value pair is a pair of linked data items: a key and a value, where the key is used as a unique identifier for the value, as shown in Fig.~\ref{FUNSD_showcase}. Automatically extracting key-value pairs is crucial for improving the efficiency of business document processing and reducing human labor. However, due to the diverse contents and complex layouts of form-like documents, key-value pair extraction from form-like document images is a challenging task. Although many deep learning based key-value pair extraction approaches \cite{jaume2019funsd, davis2019deep, xu2021layoutxlm, davis2021visual, zhang2021entity, dang2021end, wang2022LiLT} have emerged and significantly outperformed traditional rule or handcrafted feature based methods \cite{ watanabe1995layout, seki2007information, hirayama2011development} in terms of both accuracy and capability, their performance is still far from satisfactory, e.g., the current state-of-the-art F1 score achieved on the widely used FUNSD dataset \cite{jaume2019funsd} is only 75.0\% \cite{dang2021end}.

Recently, there is a trend to unify different NLP, CV, and vision-language tasks under a same framework by casting them as a language modeling or question-answering (QA) problem and training a unified model to achieve state-of-the-art results on different tasks. The same trend can also be seen in the visual document understanding (VDU) area. A recent work TILT \cite{powalski2021going} has demonstrated that a unified encoder-decoder based QA model, pre-trained on a set of labeled datasets with different annotations from different document understanding tasks, can be finetuned to achieve superior results on several visual document understanding tasks, including key information extraction \cite{huangSROIEicdar2019}, document classification \cite{harley2015RVLCDIP}, and document VQA \cite{mathew2021docvqa}. This unified QA model has three advantages: 1) The task specific design is no longer needed; 2) Diverse datasets with different annotations can be directly used to train one model; 3) The supervised multi-task pre-training algorithm can leverage information learned by one task to benefit the training of other tasks. Despite these advantages, how to extend this unified QA model to deal with many challenging relation prediction problems (e.g., key-value pair extraction) in VDU area is still unexplored. 

In this paper, we explore how to extend the QA framework to solve the key-value pair extraction problem. To this end, we propose a new Transformer-based encoder-decoder model, namely \Ours{}, to extract key-value pairs from various form-like documents. Specifically, given an input form-like document image with provided semantic entities, \Ours{} first identifies key entities from all entities, then takes these key entities as questions and feeds them into a DETR-style \cite{carion2020end} Transformer decoder to predict their corresponding answers, i.e. value entities. Different from the conventional Transformer decoder used in machine translation \cite{vaswani2017attention}, the DETR-style decoder excludes the masked attention mechanism and models the input questions in a bidirectional manner so that all the corresponding answers can be predicted in parallel. To achieve higher answer prediction accuracy, we propose a coarse-to-fine answer prediction algorithm further, which extracts multiple candidate answers for each identified question in the coarse stage and then selects the most likely one among these candidates in the fine stage. Moreover, we introduce a new spatial compatibility attention bias into the self-attention/cross-attention layers in Transformer encoder and decoder to better model the spatial relationships in each pair of entities. Thanks to these effective techniques, our \Ours{} achieves state-of-the-art results on FUNSD \cite{jaume2019funsd} and XFUND \cite{xu2021layoutxlm} datasets. The main contributions of this paper can be summarized as follows:
\begin{itemize}
    \item We formulate key-value pair extraction as a QA problem and propose a new Transformer-based encoder-decoder model, called KVPFormer, for key-value pair extraction.
    \item We introduce three effective techniques, namely a DETR-style decoder, a coarse-to-fine answer prediction algorithm and a new spatial compatibility attention bias into the self-attention/cross-attention mechanism, to improve significantly both the efficiency and accuracy of \Ours{} on key-value pair extraction tasks. 
    \item Our proposed \Ours{} achieves state-of-the-art performance on FUNSD and XFUND datasets, outperforming the previous best-performing method by 7.2\% and 13.2\% in F1 score, respectively.
\end{itemize}

\begin{figure*}[!t]
    \centering 
    \includegraphics[width=1\textwidth]{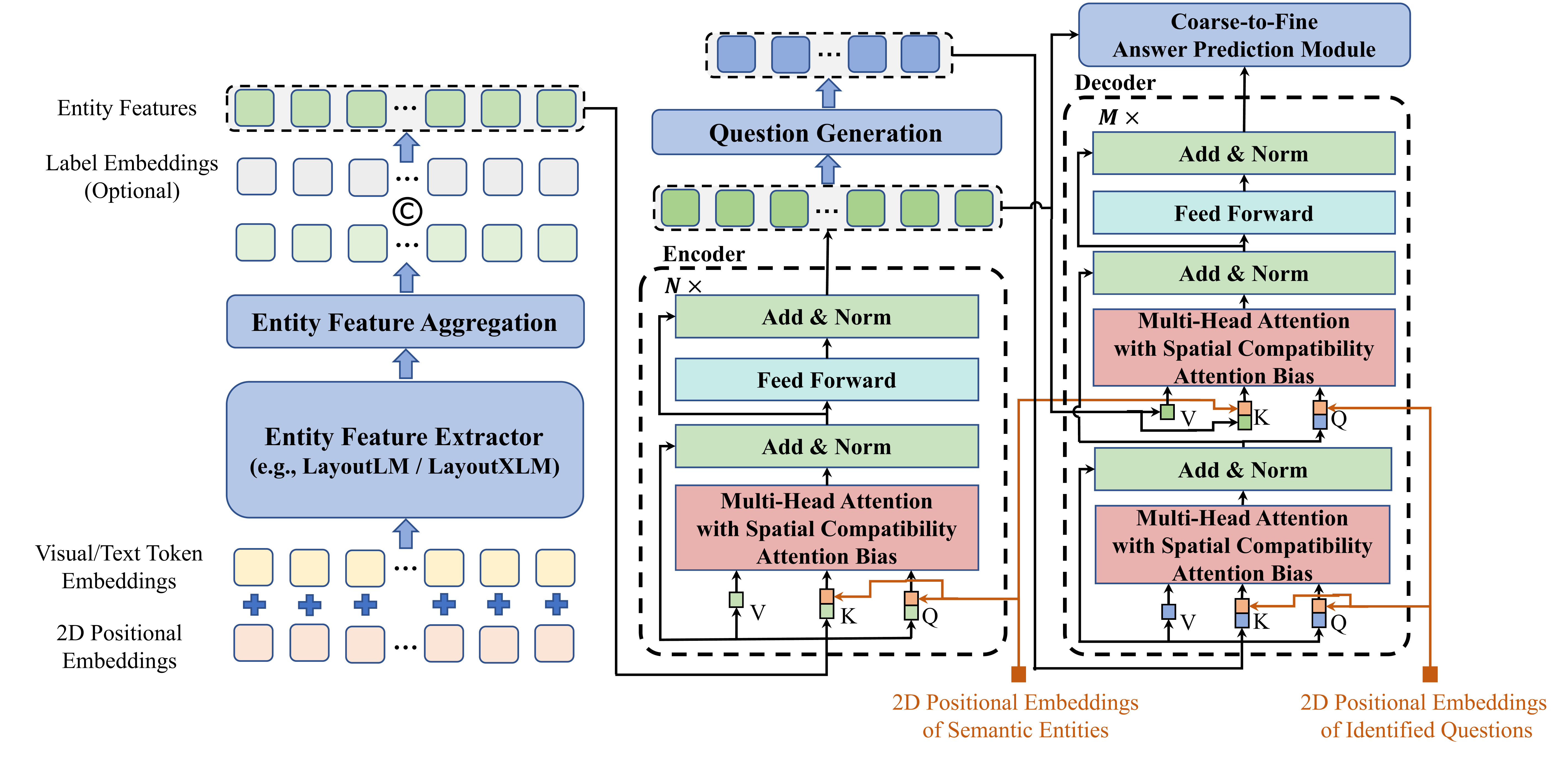}
    \caption{The architecture of \Ours{}.}
    \label{fig:overall_architecture}
\end{figure*}

\section{Related Work}
Key-value pair extraction from form-like documents has been studied for decades. Some early works, e.g., \cite{watanabe1995layout, seki2007information, hirayama2011development}, are mainly based on heuristic rules or handcrafted features in known templates, therefore they usually fail on unseen templates and are not scalable in practice. In recent years, many deep learning based key-value pair extraction approaches have emerged and outperformed these traditional methods significantly in terms of both accuracy and capability. These methods typically leverage existing text or document representation models, e.g., \cite{devlin2018bert, xu2020layoutlm, garncarek2021lambert, yu2021pick, lin2021vibertgrid}, to extract the representations of entities first, and then feed these entity representations into different key-value pair extraction models to obtain key-value pairs. Most existing key-value pair extraction methods, e.g.,  \cite{jaume2019funsd, hwang2021spatial, 9412669,10.1007/978-3-030-86159-9_27, wang2020docstruct, hong2020bros, xu2021layoutxlm, li2021structext, wang2022LiLT, gu2022xylayoutlm}, directly concatenated the representations of each entity pair and adopt a classifier like multilayer perceptron (MLP) to predict whether these two entities form a key-value pair or not. Other than this group of methods, there are some works trying to improve the accuracy of key-value pair extraction from other perspectives. For instance, \cite{davis2019deep} first predicted both the score of each candidate pair and the number of relationships belonging to each entity, then used a global optimization to derive the final set of relationships. FUDGE \cite{davis2021visual} dynamically edited the graph structure built from a document by merging text segments (graph vertices) and pruning edges in an iterative fashion to obtain the final entities and relationships. SERA \cite{zhang2021entity} adapted the biaffine parser, which is a popular dependency parsing model, to the key-value pair extraction task. MSAU-PAF \cite{dang2021end} extended the MSAU \cite{dang2021endbmvc} architecture with link-prediction between entities by integrating the PIF-PAF mechanism, Corner Pooling and Coordinate Convolution for exploiting spatial correlation.
\cite{Villota2021TextCM} concatenated the original textual content of each key entity with the textual content of the candidate value entity, and used a text classification model to determine if that combination of key and value is correct or not. 

Recently, \cite{powalski2021going} proposed to unify several document understanding tasks, including document classification, key information extraction and document VQA, through an encoder-decoder Transformer-based QA model. Although this framework achieved very promising results on the above document understanding tasks, it can not be directly applied to the key-value pair extraction task. SimpleDLM \cite{gao2021value} leveraged the idea of QA to extract key-value pairs by formulating the problem as value retrieval with arbitrary queries. However, SimpleDLM has two limitations: 1) It is not an end-to-end approach since this work assumes that the arbitrary queries (questions) have been given by users; 2) The performance of SimpleDLM is far from satisfactory (only 60.7\% in F1-score under its used evaluation metric). In this paper, we explore how to design an end-to-end QA-based approach to effectively addressing the key-value pair extraction problem so that the QA-based paradigm becomes
more general on VDU tasks.

\section{Problem Formulation}
\label{sec:taskdefinition}
Given a form-like document image $D$ composed of $n$ semantic entities $D=[E_1, E_2, ..., E_n]$, each entity $E_i$ comprises a group of words $[w_i^1, w_i^2,...w_i^m]$ and the coordinates of its bounding box $[x_i^1, y_i^1, x_i^2, y_i^2]$, where $(x_i^1,y_i^1)$ and $(x_i^2,y_i^2)$ represent the coordinates of its top-left and bottom-right corners respectively. $E_i \rightarrow E_j$ denotes that there exists a directed key-value relationship pointing from a key entity $E_i$ to a value entity $E_j$. The goal of key-value pair extraction is to extract all these key-value pair relationships from a form-like document image. In this work, we formulate key-value pair extraction as a QA problem. A new Transformer-based encoder-decoder model is proposed to first identify key entities from all entities with the encoder, then take the identified key entities as questions and feed their representations into the decoder to predict their corresponding value entities in parallel. Following the previous work \cite{zhang2021entity}, we assume that the bounding boxes and textual contents of entities have already been given.

\section{Methodology}
\label{sec:methodology}
Our proposed \Ours{} is composed of three key components: 1) An entity representation extraction module to extract a vector representation (embedding) for each entity; 2) A Transformer encoder based question identification module to identify key entities as questions from all entities; 3) A Transformer decoder based coarse-to-fine answer prediction module to predict the corresponding value entity for each identified key entity. A schematic view of our \Ours{} architecture is depicted in Figure~\ref{fig:overall_architecture}. Its details are described in the following subsections.

\subsection{Entity Representation Extraction}
 We leverage the widely used LayoutLM \cite{xu2020layoutlm} and LayoutXLM \cite{xu2021layoutxlm} models, which are both pre-trained with large amounts of document images, as the feature extractors to extract the representation of each entity on English-only and multilingual datasets, respectively. Specifically, we first serialize all the words in a document image into a 1D sequence by reading them in a top-left to bottom-right order and tokenize the word sequence into a sub-word token sequence, which is then fed into the feature extractor to get the embedding of each token. After that, we average the embeddings of all the tokens in each entity to obtain its entity representation. 

For evaluating the upper bound performance of our approach, following SERA \cite{zhang2021entity}, we further concatenate the entity label embeddings (gold labels) to entity representations:
\begin{equation}
    \boldsymbol{c}_i = \boldsymbol{e}_i \oplus \boldsymbol{l}_i,
\end{equation}
where $\boldsymbol{l}_i$ is a learnable entity label embedding, $\boldsymbol{e}_i$ is the entity representation obtained from the entity feature extractor.

\subsection{Question Identification with Transformer Encoder}
 We feed all entity representations into a Transformer encoder to identify questions. To explicitly model the spatial relationships between different entities, we introduce a new spatial-aware self-attention mechanism into the self-attention layers in the vanilla Transformer encoder. Specifically, we treat the representations of entities input into each encoder layer as content embeddings and concatenate them to their corresponding 2D positional embeddings \cite{xu2020layoutlm} to obtain the key and query embeddings in each multi-head self-attention layer. Moreover, we introduce a spatial compatibility attention bias into the self-attention mechanism to model the spatial compatibility relationship between each pair of entities. Formally, the attention score between a query $q_i$ and a key $k_j$ in a self-attention layer is denoted as:
\begin{equation}
\begin{aligned}
    \alpha_{ij} &= (\textbf{c}_{q_i} \oplus \textbf{p}_{q_i})^T(\textbf{c}_{k_j} \oplus \textbf{p}_{k_j}) + \textbf{FFN}(\textbf{r}_{q_i , k_j})  \\
    &= \textbf{c}_{q_i}^T \textbf{c}_{k_j} + \textbf{p}_{q_i}^T \textbf{p}_{k_j} + \textbf{FFN}(\textbf{r}_{q_i , k_j}),
\end{aligned}
\end{equation}
where $\textbf{c}_{q_i}$/$\textbf{c}_{k_j}$ and $\textbf{p}_{q_i}$/$\textbf{p}_{k_j}$ are the content and 2D positional embeddings of query $q_i$ / key $k_j$, respectively. $\oplus$ denotes vector concatenation. $\textbf{FFN}(\textbf{r}_{q_i, k_j})$ is the spatial compatibility attention bias where \textbf{FFN} represents a two-layer feedforward network 
 and $\textbf{r}_{q_i,k_j}$ is a spatial compatibility feature vector, which is a concatenation of three 6-d vectors:
\begin{equation}
\label{rel2d_1}
\textbf{r}_{q_i,k_j} = (\Delta(B_i, B_j), \Delta(B_i, U_{ij}), \Delta(B_j, U_{ij})),
\end{equation}
where $B_i$ and $B_j$ are the bounding boxes of $q_i$ and $k_j$, respectively; $U_{ij}$ is the union bounding box of $B_i$ and $B_j$; $\Delta(.,.)$ represents the box delta between any two bounding boxes. Taking $\Delta(B_i, B_j)$ as an example, $\Delta(B_i, B_j) = ( t^{x_{\text{ctr}}}_{ij}, t^{y_{\text{ctr}}}_{ij}, t^w_{ij}, t^h_{ij}, t^{x_{\text{ctr}}}_{ji}, t^{y_{\text{ctr}}}_{ji})$, where each dimension is given by:
\begin{equation}
\begin{aligned}
     t^{x_{\text{ctr}}}_{ij} &= (x^{\text{ctr}}_{B_i} - x^{\text{ctr}}_{B_j})/w_{B_i}, &t^{y_{\text{ctr}}}_{ij} &=  (y^{\text{ctr}}_{B_i} - y^{\text{ctr}}_{B_j})/h_{B_i}, \\
     t^w_{ij} &= \log (w_{B_i}/w_{B_j}), &t^h_{ij} &= \log (h_{B_i}/h_{B_j}), \\
     t^{x_{\text{ctr}}}_{ji} &= (x^{\text{ctr}}_{B_j} - x^{\text{ctr}}_{B_i})/w_{B_j}, &t^{y_{\text{ctr}}}_{ji} &=  (y^{\text{ctr}}_{B_j} - y^{\text{ctr}}_{B_i})/h_{B_j},
\end{aligned}
\end{equation}
where $(x^{\text{ctr}}_{B_i}, y^{\text{ctr}}_{B_i}), w_{B_i}, h_{B_i}$ and $(x^{\text{ctr}}_{B_j}, y^{\text{ctr}}_{B_j}), w_{B_j}, h_{B_j}$ are the center coordinates, width and height of $B_i$ and $B_j$, respectively.

Each feature vector output by a self-attention layer is the weighted sum of the content embeddings of all values with respect to the normalized attention score:
\begin{equation}
    \textbf{z}_i = \sum_j \frac{\exp(\alpha_{ij})}{\sum_k \exp(\alpha_{ik})} \textbf{c}_{v_j}.
\end{equation}

\subsubsection{Question Identification.}
Given the enhanced entity representations output by the Transformer encoder, we add a linear classifier to identify key entities as questions from all the entities. Essentially, question identification is a binary classification task that determines whether a given entity is a key entity or not. Note that for popular form understanding datasets like FUNSD \cite{jaume2019funsd} and XFUND \cite{xu2021layoutxlm}, in addition to key-value relationship labels, they also provide semantic labels for each entity, such as “Question”, “Answer”, “Header” and “Other”. We observe that there are no key-value relationships between “Non-other” (i.e., Question”, “Answer” and “Header”) entities and “Other” entities. Therefore, we perform question identification by predicting the label of each entity, and we keep all predicted “Non-other” entities (i.e., key entities) as questions and use the corresponding entity representations as question representations.

\subsection{Coarse-to-Fine Answer Prediction with Transformer Decoder}
\label{Coarse-to-Fine}
We feed all question representations into a Transformer decoder to predict their corresponding answers in parallel. In each decoder layer, a multi-head self-attention module is first used to model the interactions between questions and then a multi-head cross-attention module is used to model the interactions between questions and all the entities. Different from the conventional Transformer decoder that is an autoregressive model in previous works, e.g.,  \cite{vaswani2017attention,powalski2021going}, a DETR-style \cite{carion2020end} decoder is adopted here, which excludes the masked attention mechanism and models the input questions in a bidirectional manner so that all the corresponding answers can be predicted in parallel. Moreover, we also treat question representations as content embeddings and concatenate them to the 2D positional embeddings, and we add the spatial compatibility attention bias into all self-attention and cross-attention modules in the decoder, which is shown in Figure \ref{fig:overall_architecture}.

\subsubsection{Coarse-to-Fine Answer Prediction.}
We propose a coarse-to-fine answer prediction algorithm to improve the answer prediction accuracy. Specifically, for each identified question, this module first extracts multiple answer candidates in the coarse stage and then selects the most likely one among these candidates in the fine stage.

Formally, let $\textbf{H}=[\textbf{h}_1, \textbf{h}_2, ..., \textbf{h}_N]$ denote the entity representations output by the encoder and $\textbf{Q}=[\textbf{q}_1, \textbf{q}_2, ..., \textbf{q}_M]$ denote the question representations output by the decoder, where $N$ is the number of entities and $M$ is the number of identified questions in the input document image. In the coarse stage, we use a binary classifier to calculate a score $s^{\text{coarse}}_{ij}$ to estimate how likely an entity $\textbf{h}_j$ is the answer (i.e., value entities) of a question $\textbf{q}_i$ as follows:
\begin{equation}
    \label{coarse_score}
    s^{\text{coarse}}_{ij} = \text{Sigmoid}(\text{MLP}_{\text{coarse}}(\textbf{q}_i + \textbf{h}_j + \textbf{FFN}(\textbf{r}_{ij})))
\end{equation}
where $\textbf{r}_{ij}$ is the 18-d spatial compatibility feature between $\textbf{q}_i$ and $\textbf{h}_j$ described in the previous subsection; $\textbf{FFN}$ is a feedforward network that matches the dimension of $\textbf{r}_{ij}$ to that of $\textbf{q}_i$ and $\textbf{h}_j$. For each identified question $\textbf{q}_i$, we sort the scores $[s_{ij}, j=1,2,...,N]$ in a descending order and select $K$ ($K$=5) answer candidates $[\textbf{h}_{i_k}, k=1,2,...,K]$ with the top-$K$ highest scores for the fine stage.

In the fine stage, we use a multi-class (i.e., $K$-class) classifier to calculate the score $s_{i{i_k}}^{\text{fine}}$ between question $\textbf{q}_i$ and its answer candidate $\textbf{h}_{i_k}$ extracted in the coarse stage:
\begin{equation}
\label{refinement_score}
\begin{aligned}
    t_{i{i_k}} &= \text{MLP}_{\text{fine}}(\textbf{q}_i + \textbf{h}_{i_k} + \textbf{FFN}(\textbf{r}_{i{i_k}})), \\
    s_{i{i_k}}^{\text{fine}} &= \frac{\exp(t_{i{i_k}})}{\sum_k \exp(t_{i{i_k}})}.
\end{aligned}
\end{equation}
We select the highest score from scores $[s_{i{i_k}}^{\text{fine}}, k=1,2,...,K]$ and output the corresponding entity as the final answer of question $\textbf{q}_i$.

\subsection{Loss Function}
\subsubsection{Question Identification Loss.} 
We perform question identification by classifying the labels of semantic entities, whose loss function is a softmax cross-entropy loss:
\begin{equation}
    \mathcal{L}_{Q} = \frac{1}{N} \sum_i \text{CE}(\bm{p}_i, \bm{p}_i^*),
\end{equation}
where $p_i$ is the predicted result of the $i^{th}$ entity output by a Softmax function and $p_i^*$ is the corresponding ground-truth label.
\subsubsection{Coarse-to-Fine Answer Prediction Loss.} In the coarse stage of answer prediction, we adopt a binary cross-entropy loss as follows:
\begin{equation}
\begin{aligned}
    \mathcal{L}_{A}^{\text{coarse}} = \frac{1}{MN} \sum_{ij} \text{BCE}(s^{\text{coarse}}_{ij}, {s^{{\text{coarse}}^{*}}_{ij}}),
\end{aligned}
\end{equation}
where $s^{\text{coarse}}_{ij}$ is the predicted score calculated by Eq.~\ref{coarse_score} and ${s^{{\text{coarse}}^{*}}_{ij}} \in \{0,1\}$ is the ground-truth label. ${s^{{\text{coarse}}^{*}}_{ij}}$ is 1 only if the entity $\textbf{h}_j$ is the answer of question $\textbf{q}_i$, otherwise it is 0.

In the fine stage of answer prediction, we adopt a softmax cross-entropy loss as follows:
\begin{equation}
    \mathcal{L}_{A}^{\text{fine}} = \frac{1}{M} \sum_{i} \text{CE}(\bm{s}^{\text{fine}}_{i}, {\bm{s}^{{\text{fine}}^{*}}_{i}}),
\end{equation}
where $\bm{s}^{\text{fine}}_{i}=[s_{i{i_1}}^{\text{fine}},s_{i{i_2}}^{\text{fine}},...,s_{i{i_K}}^{\text{fine}}]$ are the predicted scores calculated by Eq.~\ref{refinement_score} and ${\bm{s}^{{\text{fine}}^{*}}_{i}}$ are the target labels.

\subsubsection{Overall Loss.}
All the modules in our approach are jointly trained in an end-to-end manner. The overall loss is the sum of $\mathcal{L}_{Q}$, $\mathcal{L}_{A}^{\text{coarse}}$ and $\mathcal{L}_{A}^{\text{fine}}$:
\begin{equation}
    \mathcal{L} = \mathcal{L}_{Q} + \mathcal{L}_{A}^{\text{coarse}} + \mathcal{L}_{A}^{\text{fine}}.
\end{equation}

\section{Experiments}

\subsection{Datasets and Evaluation Protocols}

We conduct experiments on two popular benchmark datasets, i.e., FUNSD \cite{jaume2019funsd} and XFUND \cite{xu2021layoutxlm}, to verify the effectiveness of our proposed approach. \textbf{FUNSD} is an English form understanding dataset which contains 149 training samples and 50 testing samples. \textbf{XFUND} is a multi-lingual form understanding dataset which contains 1393 forms in seven languages, including Chinese, Japanese, Spanish, French, Italian, German and Portuguese. For each language, there are 149 samples for training and 50 samples for testing. In this dataset, we follow two experimental settings designed in \cite{xu2021layoutxlm}, i.e., multi-task learning and zero-shot transfer learning. These two datasets can be used for many tasks, where we focus on the task of key-value pair extraction (aka entity linking on FUNSD) in this paper. The standard relation-level precision, recall and F1-score are used as evaluation metrics.

\subsection{Implementation Details}
We implement our approach based on Pytorch v1.8.0 and all experiments are conducted on a workstation with 8 Nvidia Tesla V100 GPUs (32 GB memory). Note that on FUNSD and XFUND datasets, since one key entity may have multiple corresponding value entities while each value entity only has at most one key entity, we follow \cite{zhang2021entity} to first identify all the value entities as questions and then predict their corresponding key entities as answers to obtain key-value pairs. We use a 3-layer Transformer encoder and a 3-layer Transformer decoder, where the head number, dimension of content/2D positional embedding and dimension of feedforward networks are set as 12, 768 and 2048, respectively. In training, the parameters of the entity feature extractor are initialized with well-pretrained LayoutLM$_{\mathrm{BASE}}$ for FUNSD and LayoutXLM$_{\mathrm{BASE}}$ for XFUND respectively, while the parameters of the newly added layers are initialized by using random weights with a Gaussian distribution of mean 0 and standard deviation 0.01. The models are optimized by AdamW \cite{loshchilov2017decoupled} algorithm with batch size 16 and trained for 50 epochs on FUNSD and 20 epochs on XFUND. The learning rate is set as 2e-5 for the entity feature extractor and 5e-4 for the newly added layers. The other hyper-parameters of AdamW including betas, epsilon and weight decay are set as (0.9, 0.999), 1e-8, and 1e-2, respectively.

\begin{table}[!]
\begin{center}
\footnotesize
\begin{tabular}{l | p{0.8cm}<{\centering} p{0.8cm}<{\centering} p{0.8cm}<{\centering}} \toprule
\textbf{Methods} & \textbf{Prec.} &  \textbf{Rec.} &  \textbf{F1} \\ \midrule\midrule
\tabincell{l}{LayoutLM + MLP \\ 
(Our Implementation)} & 0.3473 & 0.25 & 0.2907 \\ \midrule 
\tabincell{l}{GNN + MLP \\ \cite{9412669}} & - & - & 0.39 \\ \midrule
\tabincell{l}{SERA \cite{zhang2021entity}} & 0.6368 & 0.6842 & 0.6596 \\ \midrule
\tabincell{l}{BROS \cite{hong2020bros}} & 0.647 & 0.7083 & 0.6763 \\ \midrule
\tabincell{l}{MSAU-PAF \cite{dang2021end}} & - & - & 0.75 \\ \midrule
\tabincell{l}{SERA + gold label \\ \cite{zhang2021entity}} & 0.7402 & 0.7777 & 0.7585 \\ \midrule
\tabincell{l}{Baseline\#1 \\
(LayoutLM + SCF)} & 0.7947 & 0.7096 & 0.7498 \\ \midrule 
\tabincell{l}{Baseline\#2 \\
(LayoutLM + SCF + DP)} & 0.8132 & 0.765 & 0.7884 \\ \midrule
\midrule
\tabincell{l}{\Ours{}} & \bf 0.8504 & \bf 0.7961 & \bf 0.8223 \\ \midrule
\tabincell{l}{\Ours{} + gold label} & \bf 0.9406 & \bf 0.8788 & \bf 0.9086 \\ \bottomrule
\end{tabular}
\end{center}
\caption{Experimental results on FUNSD dataset, where ``Prec.'' and ``Rec.'' stand for Precision and Recall; ``SCF'' and ``DP'' mean Spatial Compatibility Feature and Dependency Parsing.}
\label{tab:funsd}
\end{table}

\begin{figure*}[t]
\centering 
\includegraphics[width=0.90\textwidth, height=0.53\textwidth]{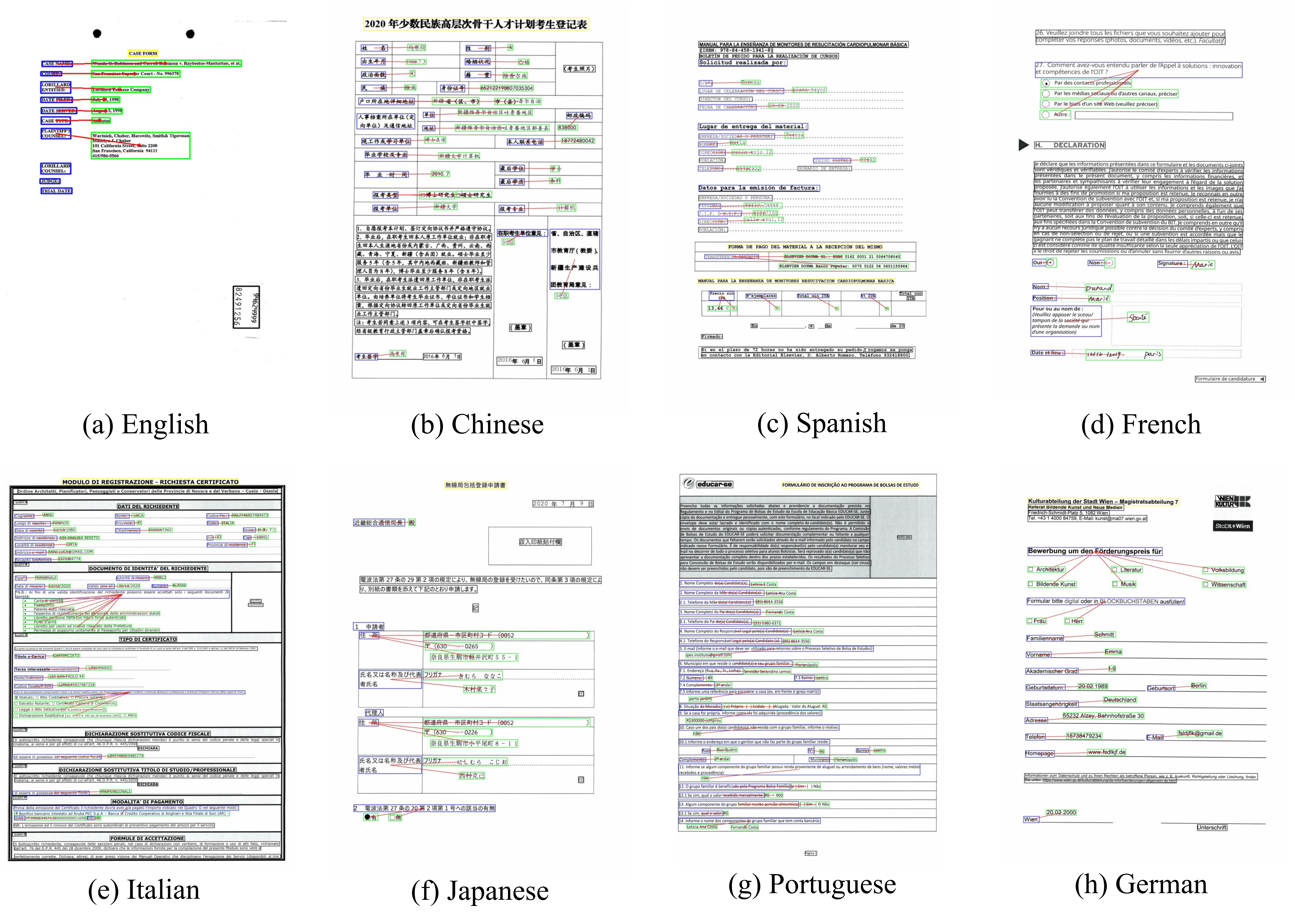}
\caption{Qualitative results of our \Ours{} on FUNSD and XFUND datasets. Bule boxes, green boxes, yellow boxes and black boxes represent “Question”, “Answer”, “Header” and “Other” entities, respectively. Red arrows stand for the predicted key-value relationships pointing from key entities to value entities. Best viewed
in color.}
\label{fig:qualitative-results}
\end{figure*}

\begin{table*}[!]
\small
\centering
\begin{tabular}{lccccccccc}
\toprule
\multicolumn{1}{c}{\bf Model} & \bf EN & \bf ZH & \bf JA & \bf ES & \bf FR & \bf IT & \bf DE & \bf PT & \bf Avg. \\ \midrule \midrule
XLM-RoBERTa$_{\mathrm{BASE}}$ &   0.3638 &  0.6797  & 0.6829 &  0.6828 &  0.6727  & 0.6937 &  0.6887  & 0.6082  & 0.6341\\
InfoXLM$_{\mathrm{BASE}}$ &    0.3699 &  0.6493  & 0.6473 &  0.6828 &  0.6831  & 0.6690 &  0.6384 &  0.5763 &  0.6145\\
LayoutXLM$_{\mathrm{BASE}}$ &    0.6671 &  0.8241 &  0.8142 &  0.8104 &  0.8221  & 0.8310 &  0.7854  & 0.7044 &  0.7823\\
LiLT[InfoXLM]$_{\mathrm{BASE}}$&  0.7407 & 0.8471 & 0.8345  & 0.8335  &  0.8466 & 0.8458  & 0.7878  & 0.7643 &   0.8125 \\
Baseline\#1 \\
(LayoutXLM$_{\mathrm{BASE}}$ + SCF) & 0.8627 &  0.8971 &  0.8486 &  0.8721 &  0.8904  & 0.8563 &  0.8302  & 0.8196 &  0.8596\\
Baseline\#2 \\
(LayoutXLM$_{\mathrm{BASE}}$ + SCF + DP)
&  0.9557 &  0.9373 &  0.9122 &  0.9349 &  0.9366  & 0.9240 &  0.9175  & 0.8874 &  0.9257\\
\midrule \midrule
\Ours{} &  \bf  0.9570 & \bf 0.9427 & \bf 0.9423 & \bf 0.9523 & \bf 0.9719  & \bf 0.9411 & \bf 0.9241  & \bf 0.9219 &  \bf 0.9442 \\ \bottomrule
\end{tabular}

\caption{Multitask learning F1-score for key-value pair extraction on XFUND dataset, where ``SCF'' means Spatial Compatibility Feature and ``DP'' stands for Dependency Parsing.}
\label{tab:xfunsd}
\end{table*}

\begin{table*}[!]
\small
\centering
\begin{tabular}{lccccccccc}
\toprule
\multicolumn{1}{c}{\bf Model} & \bf EN & \bf ZH & \bf JA & \bf ES & \bf FR & \bf IT & \bf DE & \bf PT & \bf Avg. \\ \midrule \midrule
XLM-RoBERTa$_{\mathrm{BASE}}$ &  0.2659 &0.1601& 0.2611 &0.2440 &0.2240 &0.2374 &0.2288 &0.1996& 0.2276 \\
InfoXLM$_{\mathrm{BASE}}$ &  0.2920& 0.2405 &0.2851 &0.2481 &0.2454& 0.2193 &0.2027 &0.2049 &0.2423  \\
LayoutXLM$_{\mathrm{BASE}}$ &  0.5483& 0.4494& 0.4408 &0.4708 &0.4416 &0.4090 &0.3820& 0.3685 &0.4388  \\
LiLT[InfoXLM]$_{\mathrm{BASE}}$ & 0.6276&0.4764 & 0.5081 & 0.4968 & 0.5209  &  0.4697   & 0.4169  & 0.4272 & 0.4930 \\
Baseline\#1 \\
(LayoutXLM$_{\mathrm{BASE}}$ + SCF)
&    0.8533 &  0.7633 &  0.7350 &  0.7503 &  0.7888  & 0.7572 &  0.6781  & 0.7155 &  0.7552\\
Baseline\#2 \\
(LayoutXLM$_{\mathrm{BASE}}$ + SCF + DP)
& 0.9385 &  0.9131 &  0.8864 &  0.9007 &  0.9161  & 0.8808 &  0.8771  & 0.8506 &  0.8954 \\
\midrule \midrule
\Ours{} &   \bf 0.9555 & \bf 0.9223 & \bf 0.8907 & \bf 0.9047 & \bf 0.9366  & \bf 0.8848 & \bf 0.8743  & \bf 0.8642 &  \bf 0.9041 \\ \bottomrule
\end{tabular}
\caption{Cross-lingual zero-shot transfer learning F1-score for key-value pair extraction on XFUND dataset, where ``SCF'' means Spatial Compatibility Feature and ``DP'' stands for Dependency Parsing.}
\label{tab:xfunsd-zero-shot}
\end{table*}

\subsection{Comparisons with Prior Arts}
In this section, we compare the proposed \Ours{} with several state-of-the-art methods on FUNSD and XFUND for key-value pair extraction. To better understand the superiority of our approach, we also compare our approach with two strong baselines implemented by us. In the first baseline (Baseline\#1), we follow most previous methods to directly add pair-wise entity representations output by the entity feature extractor and the corresponding spatial compatibility features together, and feed this summed representation into an MLP to predict whether there exists a key-value relationship between any two entities (referred to Eq.~\ref{coarse_score}). In the second baseline (Baseline\#2), we follow \cite{zhang2021entity} to treat relation prediction as a dependency parsing task and use softmax cross entropy loss to replace the standard binary cross entropy loss used in Baseline\#1 during optimization.

\textbf{FUNSD.} 
As shown in Table~\ref{tab:funsd}, our Baseline\#1 has already achieved the same performance as the previous best method MSAU-PAF owing to the introduction of rich spatial compatibility features. Furthermore, our Baseline\#2 has achieved a new state-of-the-art result, i.e., 78.84\% in F1-score. Compared with this strong Baseline\#2, our proposed \Ours{} is still significantly better by improving F1-score from 78.84\% to 82.23\%, which can demonstrate the advantage of our approach. To evaluate the upper bound performance of our approach, following SERA \cite{zhang2021entity}, we further concatenate the entity label embeddings (gold labels) to entity representations. The performance of our \Ours{} with gold labels can be improved to 90.86\% in F1-score. This performance gap indicates that there is still much room for improvement of our model if we can achieve better accuracy on question identification with semantic entity labels.

\textbf{XFUND.} In this dataset, we perform two experimental settings designed in \cite{xu2021layoutxlm} to verify the effectiveness of our approach: 1) Multi-task learning; 2) Zero-shot transfer learning. Moreover, following the implementation of \cite{xu2021layoutxlm}, all models here have leveraged gold labels for fair comparisons.

\textbf{1) Multi-task learning.} In this setting, all models are trained on all languages and then tested on each language. As shown in Table \ref{tab:xfunsd}, our proposed \Ours{} achieves the best result of 94.42\% in average F1-score, outperforming previous methods and our strong baselines by a substantial margin.

\textbf{2) Zero-shot transfer learning.} In this setting, all models are trained on English (i.e., FUNSD) only and tested on other languages. As shown in Table \ref{tab:xfunsd-zero-shot}, our Baseline\#1 can significantly improve the average F1-score of LayoutXLM$_{\mathrm{BASE}}$ from 43.88\% to 75.52\%, which indicates that rich spatial compatibility features play an important role in cross-lingual zero-shot transfer learning. Furthermore, \Ours{} achieves a new state-of-the-art result of 90.41\% in average F1-score, demonstrating that our approach has a better capability of transferring knowledge from the seen language to unseen languages for key-value pair extraction.

Some qualitative results of our proposed KVPFormer on FUNSD and XFUND datasets are depicted in Fig.~\ref{fig:qualitative-results}, which demonstrate that our approach can handle many challenging cases, e.g., large empty spaces between key and value entities, one key entity has multiple value entities. For failure cases, we observe that our approach cannot work equally well in extracting key-value pairs in tables, e.g., one column header is related to multiple table cells.

\begin{table}
\begin{center}
\small
\begin{tabular}{ccccc | p{0.8cm}<{\centering}} 
\toprule
\textbf{\#} & \textbf{Encoder} &  \textbf{Decoder} & \textbf{SCAB} & \textbf{C2F} & \textbf{F1} \\ \midrule\midrule
1 &  &  &  &  & 0.7498 \\ 
\midrule
2a & \checkmark &  &  &  & 0.7632  \\ 
2b & \checkmark &  & \checkmark &  & 0.7817 \\ 
\midrule
3a & & \checkmark &  &  & 0.7767\\ 
3b &  & \checkmark & \checkmark &  & 0.7906 \\ 
\midrule
4a & \checkmark & \checkmark &  &  & 0.7825\\ 
4b & \checkmark & \checkmark & \checkmark &  &  0.8017\\ 
4c & \checkmark & \checkmark & \checkmark & \checkmark &  \textbf{0.8223} \\
\bottomrule
\end{tabular}
\end{center}
\caption{Ablation studies of different component in \Ours{} on FUNSD dataset, where ``SCAB'' means Spatial Compatibility Attention Bias and ``C2F'' means Coarse-to-Fine answer prediction module.}
\label{tab:funsd-ablation}
\end{table}

\subsection{Ablation Study}
We conduct a series of ablation experiments to evaluate the effectiveness of each component in \Ours{}. All experiments are conducted on FUNSD dataset and the results are presented in Table~\ref{tab:funsd-ablation}.

\textbf{Transformer Encoder-Decoder Architecture.}
The results in \#1, \#2a-\#4a, and \#2b-\#4b rows show that 1) Both encoder and decoder can improve the performance; 2) Decoder-only is slightly better than encoder-only models; 3) The combination of encoder and decoder leads to the best performance.

\textbf{Spatial Compatibility Attention Bias.} 
As shown in results \#2a-\#2b, \#3a-\#3b, and \#4a-\#4b, no matter which architecture is used (i.e., encoder-only, decoder-only and encoder-decoder), the proposed spatial compatibility attention bias can consistently improve the performance.

\textbf{Coarse-to-Fine Answer Prediction.} 
As shown in results \#4b and \#4c, the proposed coarse-to-fine answer prediction algorithm can lead to about 2.1\% improvement in F1-score, which demonstrates its effectiveness.

\section{Conclusion and Future Work}
In this paper, we formulate key-value pair extraction as a QA problem and propose a new Transformer-based encoder-decoder model (namely KVPFormer) to extract key-value pairs, which makes the QA framework be more general on VDU tasks. Moreover, we propose three effective techniques to improve both the efficiency and accuracy of KVPFormer: 1) A DETR-style decoder to predict answers for all questions in parallel; 2) A coarse-to-fine answer prediction algorithm to improve answer prediction accuracy; 3) A new spatial compatibility attention bias to better model the spatial relationships in each pair of entities in self-attention/cross-attention layers. Experimental results on the public FUNSD and XFUND datasets have demonstrated the effectiveness of our \Ours{} on key-value pair extraction. In future work, we will explore how to train a unified QA model to solve various VDU tasks including key-value pair extraction. Moreover, we will also explore the effectiveness of our approach on more challenging but under-researched relation prediction problems in VDU area such as hierarchical key-value pair extraction and choice group extraction.

\bibliography{aaai23}

\end{document}